\documentclass[runningheads]{llncs}
\usepackage[T1]{fontenc}
\usepackage{graphicx}
\usepackage{float}
\usepackage{booktabs}
\usepackage{caption}

\usepackage{xcolor}
\usepackage{hyperref}
\usepackage[misc]{ifsym}
\usepackage{bbding} 

\begin{document}
\title{A Refer-and-Ground Multimodal Large Language Model for Biomedicine}
%
%

\author{
    Xiaoshuang~Huang\inst{1,2}\thanks{Work performed during an internship at Baidu Inc.} \and
    Haifeng~Huang\inst{1} \and
    Lingdong~Shen\inst{3} \and
    Yehui~Yang\inst{1}$^\dagger$ \and
    Fangxin~Shang\inst{1} \and
    Junwei~Liu\inst{1} \and
    Jia~Liu\inst{1}$^{(\textrm{\Letter})}$
}
\authorrunning{Xiaoshuang Huang et al.}

\institute{
Healthcare Group, Baidu Inc, Beijing 100085, China \and
China Agricultural University, Beijing 100083, China \and
MAIS, Institute of Automation, Chinese Academy of Sciences (CASIA), Beijing 100086, China \\
\email{huangxiaoshuang@cau.edu.cn, liujia9001cc@163.com}}



\maketitle              

\def\thefootnote{${(\textrm{\Letter})}$}\footnotetext{Corresponding author.}
\def\thefootnote{$\dagger$}\footnotetext{Project Leader.}

\begin{abstract}

With the rapid development of multimodal large language models (MLLMs), especially their capabilities in visual chat through refer and ground functionalities, their significance is increasingly recognized. However, the biomedical field currently exhibits a substantial gap in this area, primarily due to the absence of a dedicated refer and ground dataset for biomedical images. To address this challenge, we devised the \textbf{Med-GRIT-270k} dataset. It comprises 270k question-and-answer pairs and spans eight distinct medical imaging modalities. Most importantly, it is the first dedicated to the biomedical domain and integrating refer and ground conversations. The key idea is to sample large-scale biomedical image-mask pairs from medical segmentation datasets and generate instruction datasets from text using chatGPT. Additionally, we introduce a \textbf{R}efer-and-Groun\textbf{D} Multimodal Large Language Model for \textbf{Bi}omedicine (\textbf{BiRD}) by using this dataset and multi-task instruction learning. Extensive experiments have corroborated the efficacy of the Med-GRIT-270k dataset and the multi-modal, fine-grained interactive capabilities of the BiRD model. This holds significant reference value for the exploration and development of intelligent biomedical assistants. The repository is at \hyperref[https://github.com/ShawnHuang497/BiRD]{https://github.com/ShawnHuang497/BiRD}

\keywords{Referring and grounding  \and Instruction dataset \and Biomedicine.}
\end{abstract}

\section{Introduction}

\begin{figure}[h]
\includegraphics[width=\textwidth]{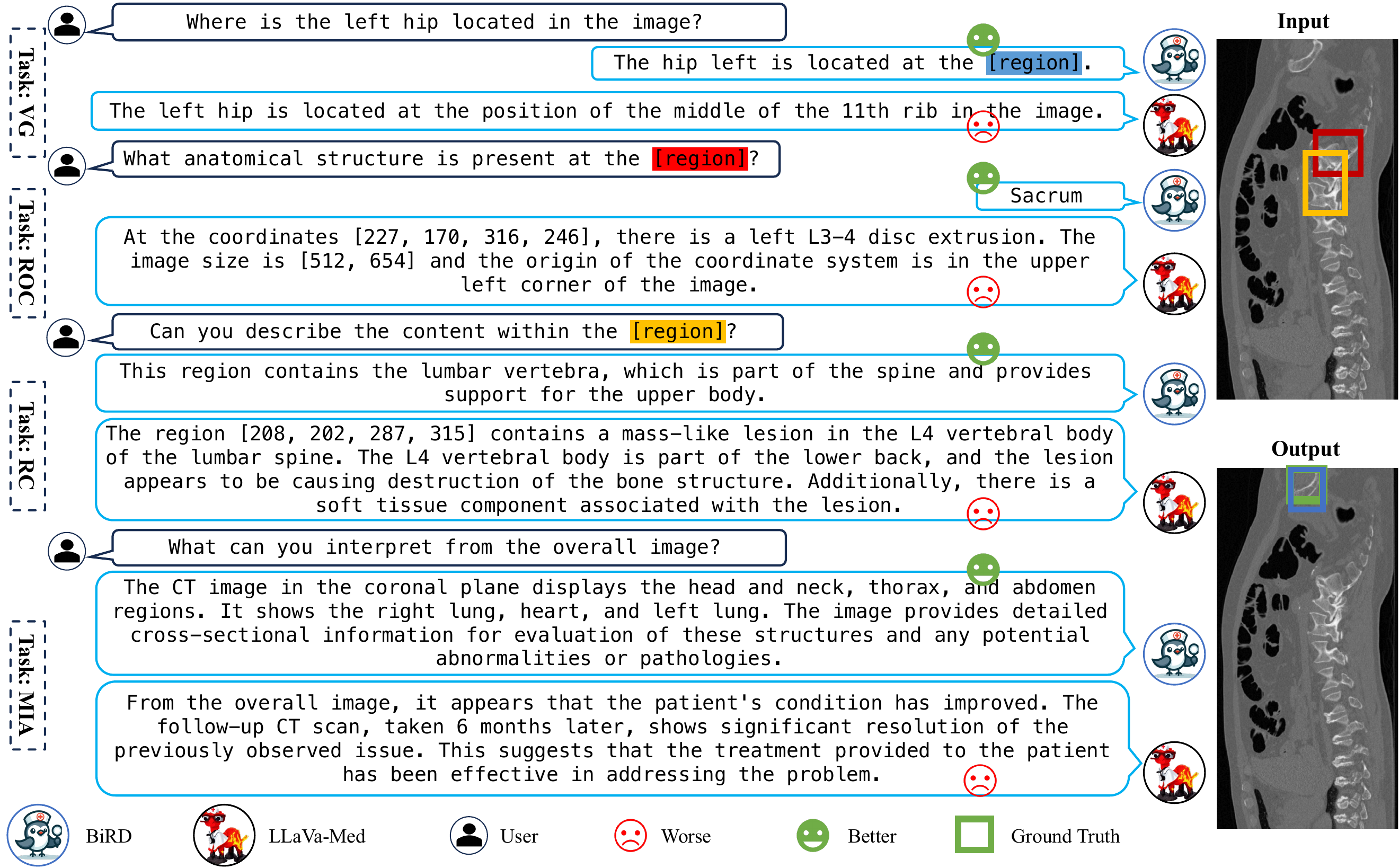}
\caption{BiRD empowers multimodal large language models in biomedicine with sophisticated referring and grounding capabilities. For more equitable comparison, we append spatial information to each LLaVa-med test, such as "The image size is [w, h], and the origin of the coordinate system is located in the upper left corner of the image.", where w and h denote width and height, respectively.} \label{fig:intro}
\end{figure}

Multimodal large language models (MLLMs) have become a popular area of research, with numerous applications in the field of visual languages, such as, Visual Question Answering (VQA), open vocabulary detection, and so on. 
Nonetheless, the unique challenges presented by the realm of biomedicine, which starkly contrasts with the natural world, often render conventional visual assistants inept. They may either refrain from responding to biomedical queries or, worse, provide inaccurate responses or entirely fabricated information~\cite{Lee_Bubeck_Petro_2023}. 

Despite existing research within the realm of biomedical MLLMs, current studies have predominantly focused on image description and VQA, leaving a notable gap in capabilities concerning referring and grounding (shown in Fig.~\ref{fig:intro}). The act of referring demands a model's accurate semantic comprehension of specified regions, while grounding necessitates the localization of regions based on semantic descriptions provided~\cite{you2023ferret}. These fine-grained multimodal capabilities are essential for both the interaction process between intelligent biomedical assistants and patients and for biomedical education. This capability not only makes the information exchange process more intuitive but also significantly enhances the accuracy and efficiency of information exchange. A key factor hindering the development of this capability in the field of biomedicine is the lack of multi-modal fine-grained interactive datasets.

To address these challenges, we develop the Bio\textbf{Med}ical \textbf{G}round-and-\textbf{R}efer \textbf{I}nstruction-Tuning (Med-GRIT-270k) dataset by leveraging the medical segmentation dataset (SA-Med2D-20M~\cite{ye2023sa}). Then a biomedical refer-and-ground multimodal large language model was explored with the Med-GRIT-270k and multi-task instruction learning method. The paper principally contributes the following: 

\begin{itemize}
\item Med-GRIT-270k Dataset. Large-scale biomedical image-mask pairs are transformed into multi-modal conversations by leveraging chatGPT~\cite{OpenAI_2023} in a novel process. It is the first dataset 
 in biomedicine to integrate referring, grounding, and conversations. 
\item The first \textbf{Bi}omedical \textbf{R}efer-and-groun\textbf{D} Multimodal Large Language Model (BiRD). It is fine-tuned by multi-task instruction learning for the biomedical domain with self-generated data. This validates the effectiveness of multi-task instruction tuning and highlights best practices for adapting the MLLMs to the specialized domain.
\item To advance biomedical multi-modal learning research, we will release the Med-GRIT-270k dataset and a comprehensive codebase for community use.
\end{itemize}

\section{Related Work}
{\bfseries Biomedical Multi-modal Large Language Models.} Amidst the rapid development of Large Language Models (LLMs) and the success of instruction-tuned LLMs within the general domain~\cite{wu2023next,liu2024visual,zhan2024anygpt,zhang2023gpt4roi,chen2023minigpt,you2023ferret}, researchers in the biomedical field have been fervently exploring the expansion of these models' capabilities. 
Recent studies have increasingly concentrated on the domain of MLLMs, with notable endeavors within the biomedical sector including BioMedGPT~\cite{luo2023biomedgpt}, RadFM~\cite{wu2023towards}, LLaVa-Med~\cite{li2024llava}, and so on~\cite{zhang2023large,liu2023medical,han2023multimodal,wang2022medclip,eslami2023pubmedclip,shen2024segicl}. 
These methodologies have significantly propelled the development of MLLMs in the biomedical realm. For instance, LLaVa-Med~\cite{li2024llava}, utilizing pre-trained LLMs for visual instruction tuning, has established a unique, end-to-end multi-modal biomedical chatbot capable of processing image inputs.
RadFM~\cite{wu2023towards} is a MLLM supporting 2D/3D radiographic imaging input for the medical domain. However, due to various challenges, biomedical MLLMs capable of supporting fine-grained interactions have yet to emerge.

{\bfseries MLLMs for Referring and Grounding.} 
In natural images, the large-scale public datasets have greatly supported the exploration into the sophisticated understanding abilities of multimodal large language models (MLLMs), such as Gpt4ROI~\cite{zhang2023gpt4roi}, Ferret~\cite{you2023ferret}, QWen-VL~\cite{bai2023qwen2}, and so on. Although some work~\cite{li2023lvit,huang2024cross} has already begun to investigate grounding in biomedicine, it can only be applied to small models, as the amount of data is limited and there are only a few modalities. The paramount factor underlying the success of these initiatives is their access to pertinent, large-scale datasets. For instance, QWen-VL uses around 80M data for referring and grounding. However, the multi-modal fine-grained interactive dataset in biomedical is virtually nonexistent.

\section{Med-GRIT-270k: Biomedical Ground-and-Refer Instruction-tuning Dataset}
We've created the first biomedical refer-and-ground instruction-tuning dataset to address the lack of such resources. It was generated through the collaborative efforts of humans and Artificial Intelligence (AI), derived from large-scale biomedical image segmentation datasets. The generation process can be divided into three steps: (i) Manually generating instance-level meta information for each image based on its mask. (ii) Employing an AI assistant to generate global information for the images. (iii) Utilizing the AI assistant to craft fine-grained conversations based on the meta information and global image information obtained in the previous steps. 

\begin{figure}[t]
\includegraphics[width=\textwidth]{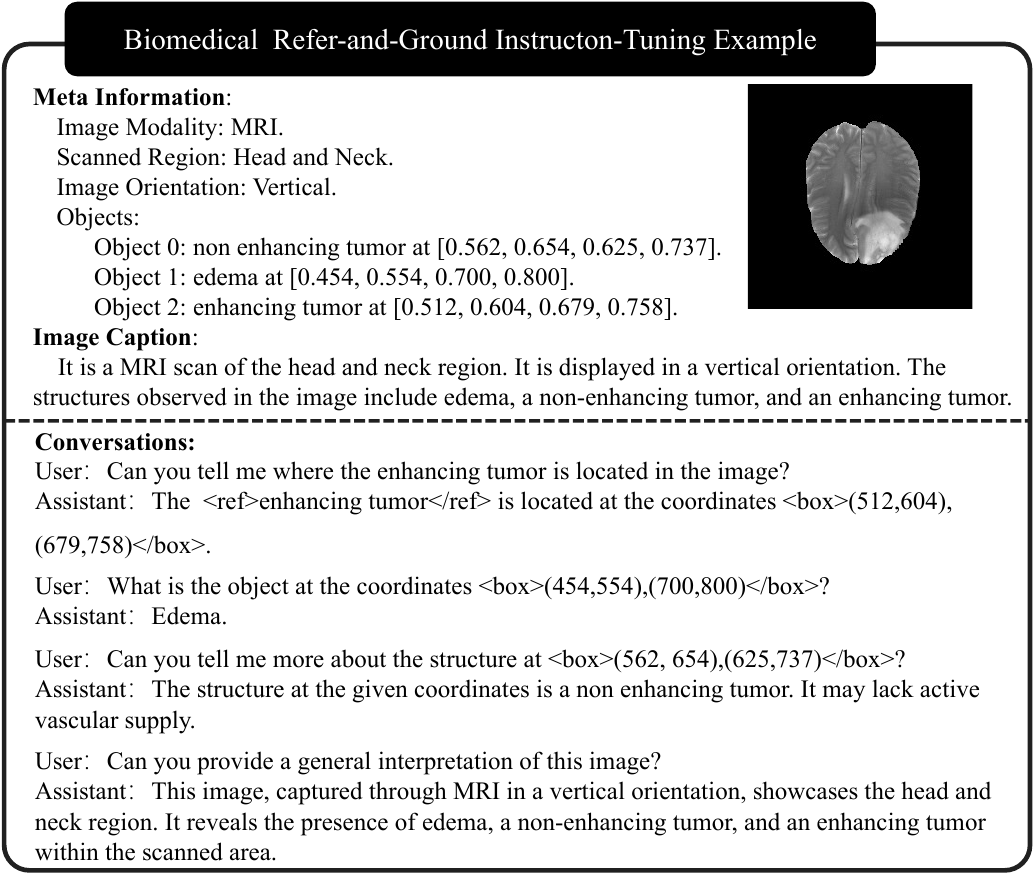}
\caption{An instance of our generated instruction-following data. Top: the meta information is created according to rules in medical segmentation datasets, and the image caption was generated from chatGPT. Bottom: the instruction following data generated by chatGPT.} \label{fig:example}
\end{figure}

\textbf{Generating Instance-level Meta Information.} We first sampled biomedical image-mask pairs from the SA-Med2D-20M~\cite{ye2023sa}. Ultimately, approximately 60K images were sampled from this dataset, considering the diversity of modality and redundancy. For instance, the original dataset includes a plethora of 2D slices from 3D data, leading to excessive data similarity. Subsequently, we calculated the coordinates of each instance based on the instance-level masks. Specifically, spatial locations are delineated via the textual representation in the format [$X_{topleft}$, $Y_{topleft}$, $X_{bottomright}$, $Y_{bottomright}$], and normalize the coordinates to fall within the range [0,1]. Finally, we enrich the images with additional details to compile the meta information, which includes modality, scanned region, orientation, and object coordinates.

\textbf{Generating Image Captions.} We utilize meticulously designed prompts along with the meta information provided to ChatGPT~\cite{OpenAI_2023}, thereby acquiring the global information for each image.

\textbf{Biomedical Instruction-Tuning Data.} Spatial understanding is manifested through various task formats. This primarily encompasses two distinct types and their corresponding task names: (i) Region-in and Text-out: Referring Object Classification (ROC), Referring Captioning (RC), (ii) Text-in and Region-out: Visual Grounding (VG), and (iii) Text-in and Text-out: Medical Image Analysis (MIA). To reduce ambiguity and enhance the model's capability for fine-grained visual comprehension, some essential strategies are adopted.
The special tokens (<ref> and </ref>) are introduced, marking the content referred to by the bounding box. This aptly associates bounding boxes with their corresponding descriptive words or sentences. Subsequently, we instructed ChatGPT to design a question and answer for each task.

Finally, We mapped the coordinates within the range [0, 1000] and reformatted them as ($X_{topleft}$, $Y_{topleft}$), ($X_{bottomright}$, $Y_{bottomright}$). To differentiate between detection strings and regular text strings, two special tokens (<box> and </box>) are appended at the start and end of the bounding box string, respectively. Fig.~\ref{fig:example} shows an example of our instruction-following data.

\section{Multi-task Instruction Learning}

\begin{figure}[ht]
\includegraphics[width=\textwidth]{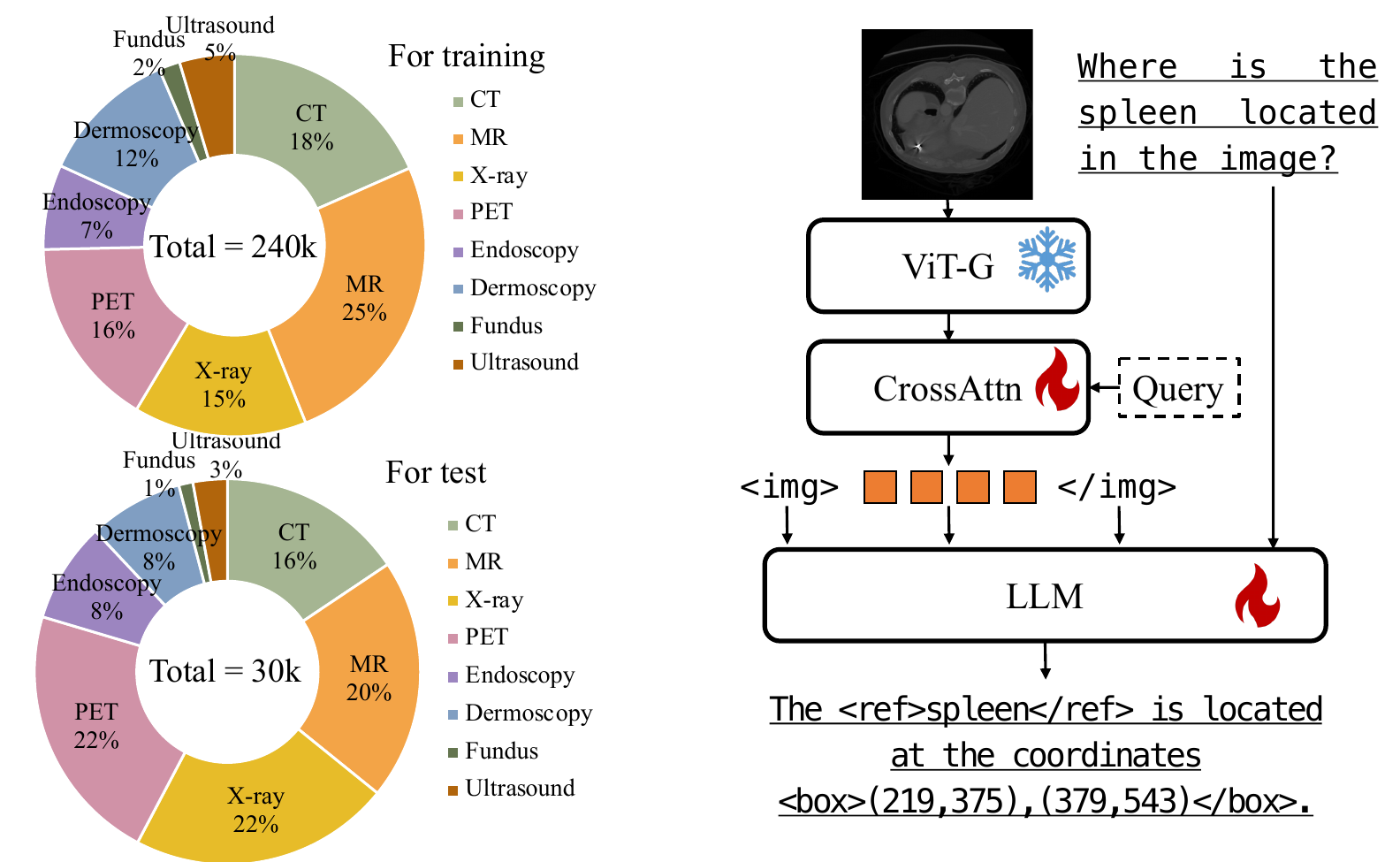}
\caption{\textbf{Overreview.} Left: the training set (Top) and test set (Bottom) distribution of conversation turns in Med-GRIT-270k we collected. Right: the architecture of the Biomedical refer-and-ground multimodal large language model (\textbf{BiRD}), which is based on Qwen-VL~\cite{bai2023qwen2}. We have developed it from the 240k data and evaluated it on 30k data.} 
\label{fig1}
\end{figure}

We aim to imbue MLLMs with grounding and referring capacities via multi-task learning, simultaneously ensuring the retention of the MLLM's essential conversational proficiency. This section will henceforth elucidate from two perspectives: the architecture of the model and multi-task instruction training.

\subsection{Model Architecture}
We utilize Qwen-VL~\cite{bai2023qwen2}, a comprehensive multimodal conversational model, as the foundational general-domain language model. Specifically, the visual encoder employs the Vision Transformer (ViT)~\cite{dosovitskiy2020image} architecture, initialized with pre-trained weights from OpenAI's CLIP ViT-BigG~\cite{ilharco10openclip}. The vision-language adapter utilizes cross-attention with a trainable query. The large language model incorporates the pre-trained Qwen-7B~\cite{bai2023qwen}.


\subsection{Multi-task Instruction Training}
Considering that the base model already possesses the capability to refer or ground within natural images, we employ only one stage to finetune it based on the pre-trained base model on the Med-GRIT-240k dataset. As illustrated in Fig.~\ref{fig1}), We solely fine-tune the cross-attention and LLM parameters, while the visual encoder remains frozen. The input images are processed through the ViT-BigG~\cite{ilharco10openclip} and vision-language adapter, yielding fixed-length sequences of visual features. We then append the markers (<img> and </img>) to the start and end of the image feature sequence, respectively, to denote the beginning and end of visual content. 
We fine-tuned the model using a dataset comprising 60k images and a total of 240k dialogue turns. The global training batch size is 128. The learning rate is $2e-5$ and the scheduler is cosine. The multi-task instruction training just took 30 hours on $4 \times A100(40G)$ GPUs.

\section{Experiments}
In this section, we execute a thorough evaluation across diverse multimodal tasks to holistically gauge our models' proficiency in visual comprehension. 

\textbf{Evaluation dataset.}
We randomly selected approximately 12\% of the images and dialogues from the constructed Med-GRIT-270k dataset to serve as the test set. Given that a single 3D dataset contains multiple data slices, we extracted cases in their entirety to prevent leakage of test set data into the training set. This ensures that different slices from the same 3D dataset do not concurrently appear in both the training and test sets, thereby guaranteeing the reliability of the test results.

\textbf{Evaluation metrics.}
The evaluation metrics for the four tasks are Recall@0.5, Recall, Spice~\cite{anderson2016spice}, and mBMR, respectively. Recall@0.5 denotes a prediction as correct only when the intersection over union (IoU) between the predicted bounding box and the ground truth exceeds 0.5. The mBMR utilized for assessing the MIA task is the mean value of BLEU@4~\cite{papineni2002bleu}, METEOR~\cite{banerjee2005meteor}, and ROUGE-L~\cite{lin2004rouge}, offering a more comprehensive evaluation of the prediction quality than a solitary metric.

\begin{table}[t]
\centering
\caption{Comparison with LLaVa-Med~\cite{li2024llava} and study on the multimodal dataset scales.}
\label{tab:ablation_data}
\resizebox{\textwidth}{!}{%
\begin{tabular}{c|c|cccc|c}
\toprule
\textbf{Model} & \textbf{Test dataset} & \textbf{VG (Recall@0.5$\uparrow$)} & \textbf{ROC (Recall$\uparrow$)} & \textbf{RC (SPICE $\uparrow$)} & \textbf{MIA (mBMR$\uparrow$)} & \textbf{Average$\uparrow$} \\ \midrule
LLaVa-Med~\cite{li2024llava}  & Med-GRIT-Test30k & 0 & 2.75 & 8.18 & 11.20 & 5.53 \\ 
BiRD-Med-GRIT-20k  & Med-GRIT-Test30k & 38.59 & 47.94 & 29.02 & 27.22 & 35.69 \\
BiRD-Med-GRIT-40k  & Med-GRIT-Test30k & 46.30 & 51.84 & 50.32 & 30.14 & 44.65 \\
BiRD-Med-GRIT-80k  & Med-GRIT-Test30k & 52.87 & 52.02 & 52.84 & 44.83 & 50.64 \\
BiRD-Med-GRIT-270k & Med-GRIT-Test30k & \textbf{53.92} & \textbf{65.33} & \textbf{55.23} & \textbf{52.17} & \textbf{56.66} \\ 
\midrule
LLaVa-Med~\cite{li2024llava}  & LLaVa-Med-qa0.2k & - & - & - & \textbf{20.04} & - \\
BiRD-Med-GRIT-270k & LLaVa-Med-qa0.2k & - & - & - & 10.55 & - \\ 
\bottomrule
\end{tabular}%
}
\end{table}

\textbf{Comparison.} As shown in Table~\ref{tab:capability}, we are the pioneers in developing a medical MLLM with referring and grounding capabilities, and existing MLLMs (such as Qwen-VL~\cite{bai2023qwen2}, GPT-4~\cite{OpenAI_2023}, MiniGPT-v2~\cite{chen2023minigpt}, etc.) have not seen medical referring and grounding data. So we will not compare them on evaluation metrics, as it would be profoundly unfair. 

\makeatletter
  \newcommand\figcaption{\def\@captype{figure}\caption}
  \newcommand\tabcaption{\def\@captype{table}\caption}
\makeatother

\begin{figure*}[t]
\begin{minipage}[]{0.55\textwidth}
\centering
\tabcaption{The capabilities of various biomedical MLLMs. Note that the "Modality" denotes image modality.}
\label{tab:capability}
\begin{tabular}{cccccc}
\toprule
\textbf{}            & \textbf{MIA}   & \textbf{ROC}   & \textbf{RC}    & \textbf{VG}    & \textbf{Modality} \\ \midrule
LLaVa-Med~\cite{li2024llava}            & \CheckmarkBold & \XSolidBrush   & \XSolidBrush   & \XSolidBrush   &     5              \\
RadFM~\cite{wu2023towards}                & \CheckmarkBold & \XSolidBrush   & \XSolidBrush   & \XSolidBrush   &     6              \\
Med-palm m~\cite{tu2024towards}           & \CheckmarkBold & \XSolidBrush   & \XSolidBrush   & \XSolidBrush   &  6         \\ \midrule
\textbf{BiRD} & \CheckmarkBold & \CheckmarkBold & \CheckmarkBold & \CheckmarkBold & \textbf{8}                 \\ \bottomrule
\end{tabular}
\end{minipage}
\hfill
\begin{minipage}[]{0.4\textwidth}
    \centering
    \includegraphics[width=\textwidth]{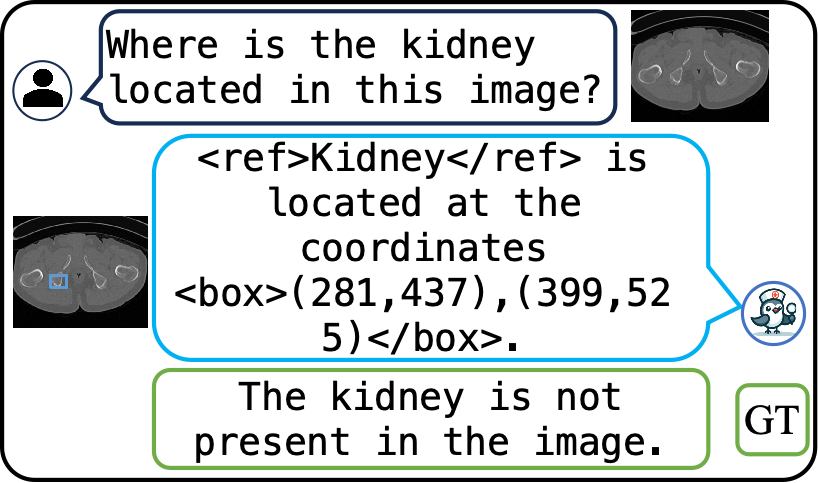}
    \caption{The example of object hallucination in BiRD.}\label{fig: hullu}
\label{fig:Hyper-parameter}
 \end{minipage}
\end{figure*}

As illustrated in Table~\ref{tab:ablation_data}, we present the quantitative test outcomes for LLaVa-Med~\cite{li2024llava} and the impact of the data scale on these results. Between rows 3 and 6, we observe the performance of the BiRD-Med-GRIT model across varying data scales. With the expansion of training data, all metrics exhibit significant enhancements, with the average rising from 35.69 to 56.66. This underscores the efficacy of augmenting dataset size in bolstering the model's proficiency on multimodal datasets. Notably, at the 240k dataset level, the model achieved the highest scores across all metrics, showcasing optimal overall performance.

From the first and sixth rows of Table~\ref{tab:ablation_data}, it is evident that the LLaVa-Med~\cite{li2024llava} model demonstrates subpar performance on the Med-GRIT-Test30k dataset, particularly in terms of no efficacy in region-level visual content localization (with the Recall@0.5 of 0).
Simultaneously, we evaluated our model on the LLaVa-Med qa-0.2k test set as well. As indicated in the last two rows of Table~\ref{tab:ablation_data}, due to not being trained on the LLaVa-Med~\cite{li2024llava} dataset, our performance metrics on its test set were marginally lower than its own. However, on similar MIA tasks within our test set, LLaVa-Med~\cite{li2024llava}(with an mBMR of 11.20), significantly underperformed in comparison to our model (with an mBMR of 52.17).



\begin{table}[t]
\centering
\caption{The performance of the BiRD model across various tasks and modalities on the Med-GRIT-270k test dataset.}
\label{tab:main_result}
\resizebox{\textwidth}{!}{%
\begin{tabular}{c|c|cccccccc|c}
\toprule
 & \textbf{Metric} & \textbf{CT} & \textbf{MR} & \textbf{X-ray} & \textbf{PET} & \textbf{Endoscopy} & \textbf{Dermoscopy} & \textbf{Fundus} & \textbf{Ultrasound} & \textbf{Average} \\ \midrule
\textbf{VG}   & Recall@0.5$\uparrow$ & 44.47 & 29.26 & 41.73 & 56.46 & 53.60 & 75.63 & 84.15 & 46.04 & 53.92 \\
\textbf{ROC}   & Recall$\uparrow$ & 34.76 & 61.79 & 53.74 &  -     & 60.40 & 96.61     & -     & 84.65     &  65.33   \\
\textbf{RC}   & Spice$\uparrow$ & 41.88 & 51.69 & 37.39 & 47.95 & 54.07 & 77.44 & 48.73 & 82.65 & 55.23    \\
\textbf{MIA}  & mBMR$\uparrow$ & 47.01 & 49.35 & 37.17 & 57.15 & 39.91 & 72.13 & 48.87 & 65.78 & 52.17 \\ \midrule
\textbf{Average} & - & 43.03 & 48.02 & 42.51 & 53.85 & 51.99 & 80.45 & 60.58 & 69.78 &  -     \\ \bottomrule
\end{tabular}%
}
\end{table}
\textbf{Main Results.} 
As shown in Table~\ref{tab:main_result}, we display the performance of the BiRD model across four distinct tasks in eight different medical imaging modalities.
The \textbf{ROC} task tests the MLLM's understanding of text related to specific image areas and their visual details. The PET and Fundus, which focus on only one category, are not trained or evaluated. We find the recall of ROC mainly depends on the variety and distinctiveness of objects and features across image modalities.
The \textbf{RC} task tests the model's ability to recognize image regions and describe them in words. The model does well with Ultrasound and Dermoscopy images but struggles with the more diverse CT images, where performance lags.
The \textbf{VG} task tests how well the model matches text descriptions to image areas. MR modality performed the worst, likely because it mostly features tumor tissues, with far fewer anatomical structures. This issue is also seen in ultrasound images.
The \textbf{MIA} task checks the model's understanding of medical images. The 4th row in Table~\ref{tab:main_result} shows the model has some level of analysis and understanding across almost all modalities.

Across the four evaluated tasks, it is apparent that the Dermoscopy modality consistently exhibits the highest performance metrics. This can be attributed to the distinct visual features, a reduced number of object categories, and the substantial proportion of the image occupied by the object regions, collectively simplifying the task for this particular modality.

\textbf{Object Hallucination.}
As Fig.~\ref{fig: hullu} shows, we have also observed instances of object hallucination in BiRD. This phenomenon is common and has also been observed in other MLLMs~\cite{li2023evaluating}. We believe this is attributed to the fact that the model's visual encoder is frozen, and its initialized parameters have scarcely encountered medical imaging, resulting in a lack of comprehensive understanding of specific domains or topics in feature extraction. In a word, this phenomenon should receive increased attention in future research endeavors.

\section{Conclusion}
In this paper, to develop a single MLLM assistant capable of handling multiple vision-language tasks, we propose a Med-GRIT-270k dataset. By leveraging the dataset, we introduce the BiRD model, a \textbf{Bi}omedical \textbf{R}efer-and-Groun\textbf{D} Multimodal Large Language Model. We verified BiRD on a diverse 30k question-and-answer test set, encompassing multimodal and multitask scenarios. The BiRD showcases a highly promising direction for developing intelligent biomedical assistants. To our knowledge, Med-GRIT-270k and BiRD are respectively the first refer-and-ground dataset and fine-grained interactive MLLM in the realm of biomedicine. We will release both the dataset and model to foster the development of intelligent biomedical assistants.

\textbf{Limitations.} Although this work developed a novel multimodal dataset in biomedicine, during the data construction process, most of the raw datasets only annotated certain organs or diseases for a sample. This makes it difficult to construct highly correlated negative samples. This issue will be a focus in the subsequent data construction work.

%
%
%
\bibliographystyle{splncs04}
%

\end{document}


%
\title{Supplementary}
%
%
%
%
%

\begin{table}[t]
\caption{Examples of task instruction following formats.}\label{tab2}
\begin{tabular}{l|ll}
\hline
\textbf{Task} &  \textbf{Two randomly chosen examples from many}\\
\hline
ROC  &  {What is the anatomical structure located at the coordinates <box>?}\\
  & {What is the object located at the coordinates <box>?}\\
\hline
RC &  {Can you describe the content within the coordinates <box>?} \\
 & {Can you describe what is seen at the coordinates <box>?}\\
\hline
VG & {Where is the <object> located in the image?} \\
 & {What are the coordinates for the given <object> in the image?}\\
\hline
MIA & {Based on the image, what can you infer about the overall medical condition?} \\
 & {How would you interpret the overall medical condition shown in the image?}\\

\hline
\end{tabular}
\end{table}




 




\begin{figure}[t]
\includegraphics[width=\textwidth]{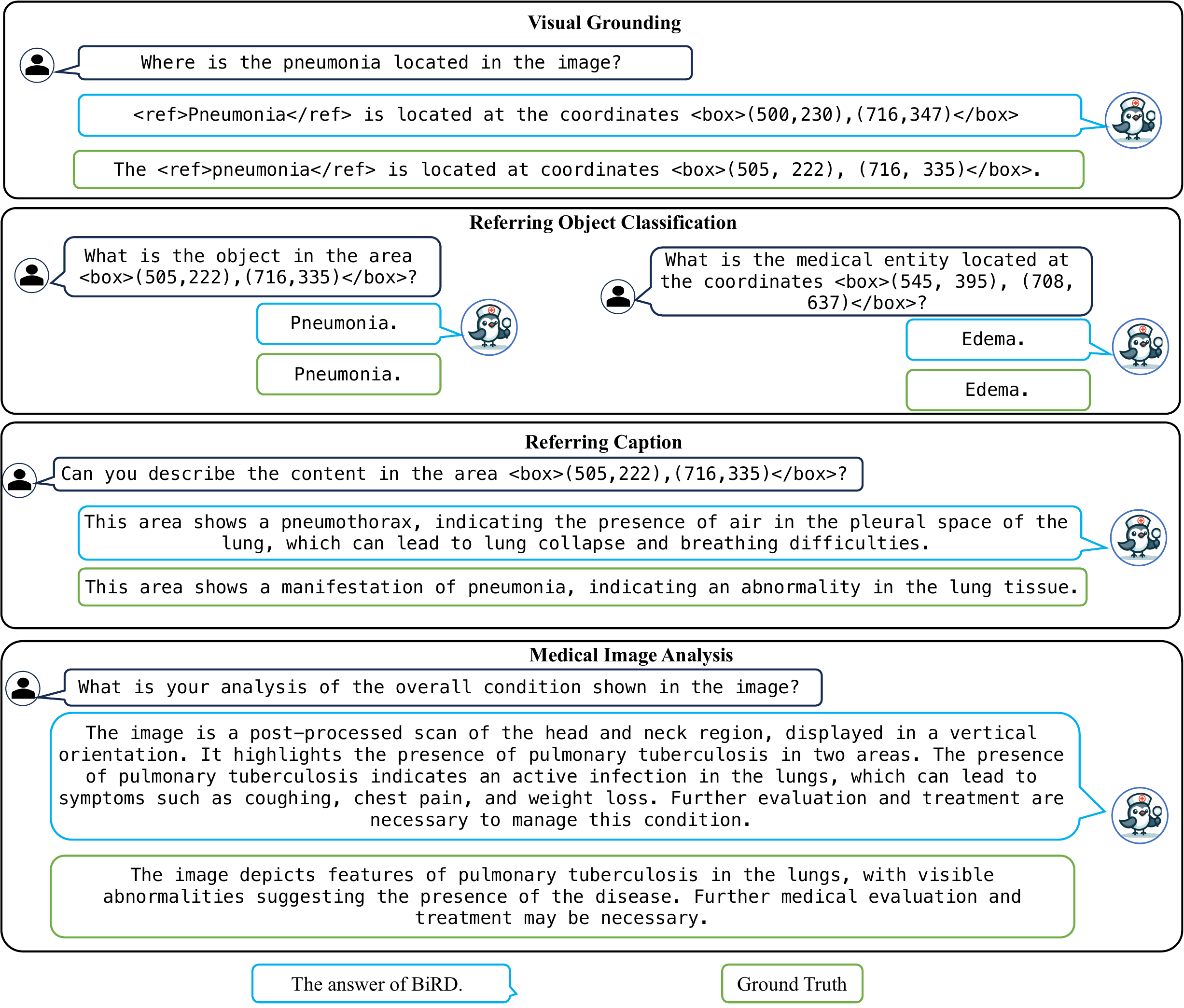}
\caption{More visualization examples of the BiRD model's performance across various tasks. Please note, due to space constraints, the actual images inputted during testing are not displayed in the figures.} \label{fig:more_case}
\end{figure}

\begin{figure}
\includegraphics[width=\textwidth]{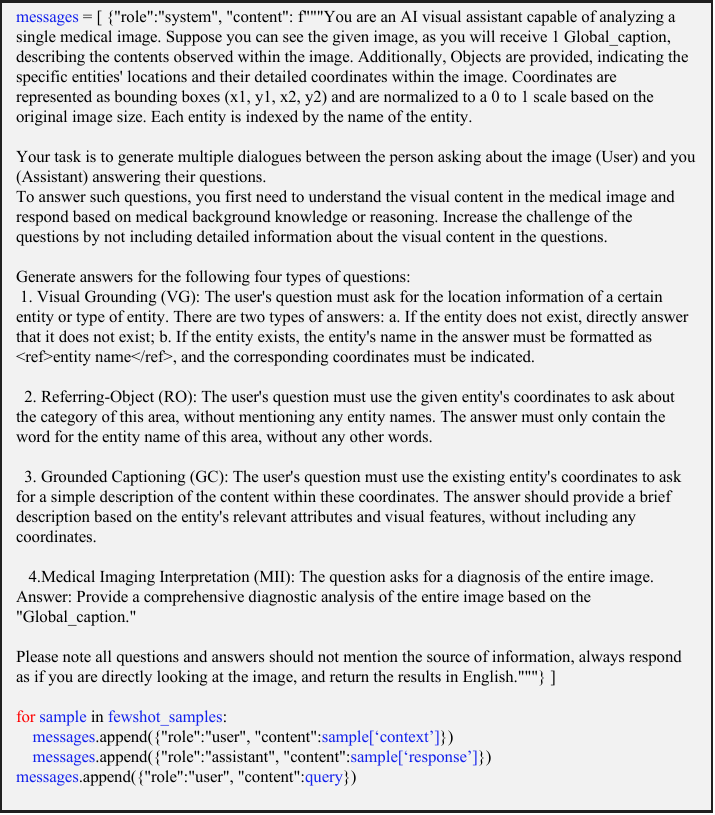}
\caption{In this example, we provide the prompt used to generate the refer-and-ground instruction tuning data.} \label{fig:prompt}
\end{figure}